\documentclass[10pt,twocolumn,letterpaper]{article}

\usepackage{iccv}              

\usepackage{amsmath,amsfonts,bm}

\def\eqref#1{equation~\ref{#1}}

\def\1{\bm{1}}

\def\vmu{{\bm{\mu}}}

\def\vgamma{{\bm{\gamma}}}
\def\vdelta{{\bm{\delta}}}

\def\vxi{{\bm{\xi}}}

\def\vc{{\bm{c}}}

\def\vp{{\bm{p}}}

\def\vv{{\bm{v}}}

\def\vx{{\bm{x}}}

\def\mB{{\bm{B}}}
\def\mC{{\bm{C}}}
\def\mD{{\bm{D}}}

\def\mM{{\bm{M}}}

\def\mO{{\bm{O}}}
\def\mP{{\bm{P}}}

\def\mV{{\bm{V}}}

\def\mX{{\bm{X}}}

\def\mSigma{{\bm{\Sigma}}}

\DeclareMathAlphabet{\mathsfit}{\encodingdefault}{\sfdefault}{m}{sl}
\SetMathAlphabet{\mathsfit}{bold}{\encodingdefault}{\sfdefault}{bx}{n}

\newcommand{\R}{\mathbb{R}}

\definecolor{iccvblue}{rgb}{0.21,0.49,0.74}

\definecolor{urlred}{rgb}{1,0.2,0.6}
\usepackage[pagebackref,breaklinks,colorlinks,allcolors=iccvblue,urlcolor=urlred]{hyperref}

\def\confName{ICCV}
\def\confYear{2025}

\usepackage{lipsum}
\usepackage{graphicx}
\usepackage{amsmath}
\usepackage{amssymb}
\usepackage{booktabs}
\usepackage{multirow}
\usepackage{bbding}
\usepackage{pifont} 
\usepackage{flushend}
\usepackage{placeins}
\usepackage[pagebackref,breaklinks,colorlinks]{hyperref}
\usepackage[capitalize]{cleveref}
\usepackage{enumitem}
\usepackage{tabularx}
\usepackage{bm}
\usepackage{algorithmic}
\usepackage{algorithm}

\definecolor{lightred}{rgb}{1, 0.5, 0.5}
\definecolor{lightblue}{rgb}{0.5, 0.6, 1}

\title{\model{}: Dynamic Urban Scene Reconstruction with \bz{} Curve Gaussian Splatting}

\author{%
  Zipei Ma$^{1,2*}$
  \quad 
  Junzhe Jiang$^{1*}$
  \quad 
  Yurui Chen$^{1}$
  \quad
  Li Zhang$^{1,2\dagger}$
  \\
  $^1$School of Data Science, Fudan University
  \quad 
  $^2$Shanghai Innovation Institute
  \vspace{.5em} 
  \\
  \href{https://github.com/fudan-zvg/BezierGS}{\texttt{github.com/fudan-zvg/BezierGS}}
}

\newcommand{\model}{\textit{B\'ezierGS}}
\newcommand{\bz}{B\'ezier}

\newcommand{\RR}{\mathbb{R}}

\newcommand{\gset}{{\mathcal{G}}}
\newcommand{\gausso}{{o}}
\newcommand{\gaussmu}{{\bm{\mu}}}
\newcommand{\gaussq}{{\mathbf{q}}}
\newcommand{\gausss}{{\bm{s}}}
\newcommand{\gaussc}{{\bm{c}}}

\begin{document}
\twocolumn[{%
\renewcommand\twocolumn[1][]{#1}%
\maketitle

\vspace{-10mm}
\begin{center}
\centering
\includegraphics[width=\linewidth]{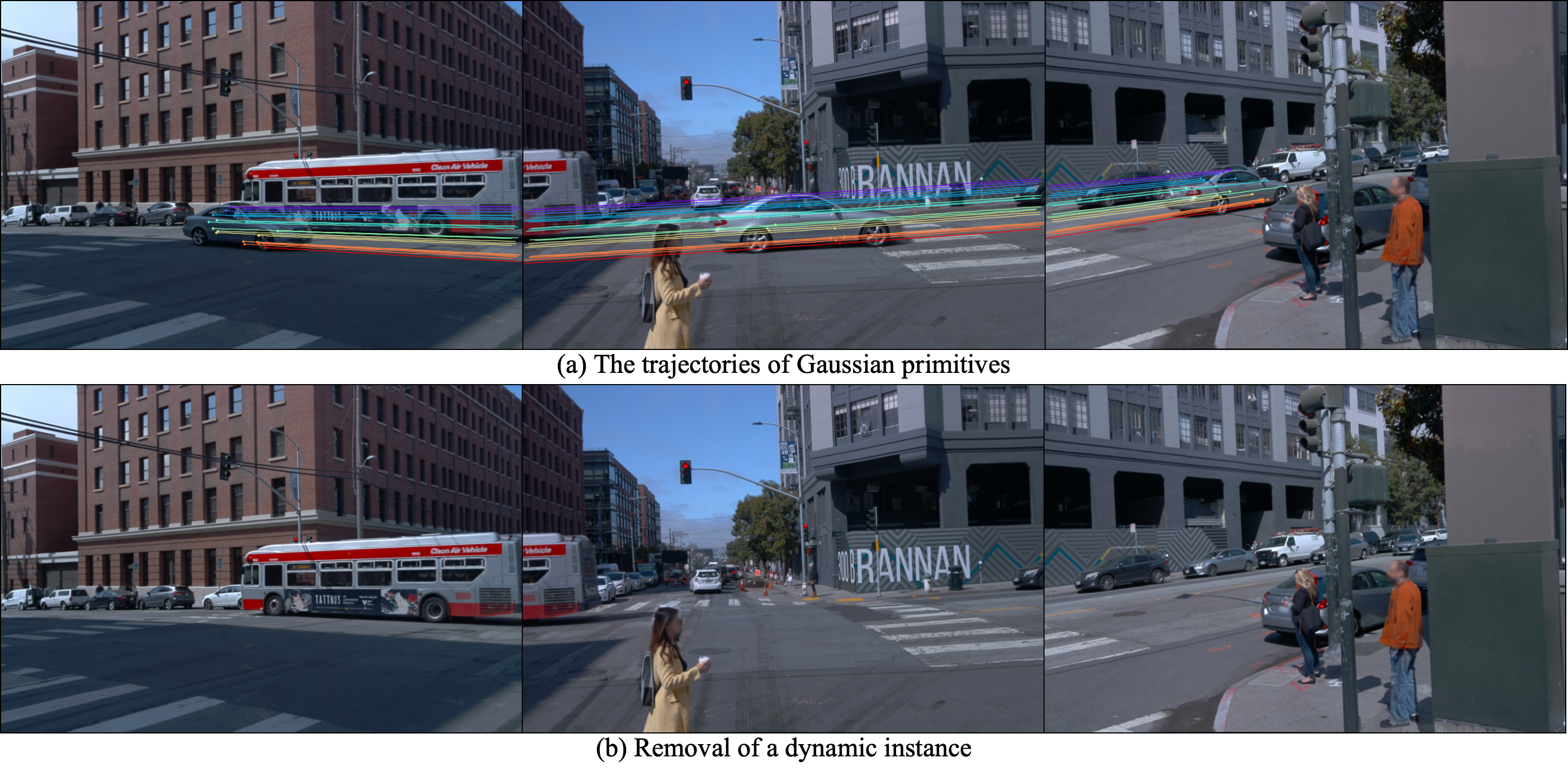}
\vspace{-9mm}
\captionof{figure}{
Our proposed \model{} effectively captures the motion of dynamic elements while accurately distinguishing between static and dynamic components in the scene.
(a) We render a dynamic vehicle instance at three different timestamps and visualize the trajectory of its Gaussian primitives, demonstrating the model’s ability to accurately represent target trajectories. (b) By leveraging our explicit modeling of dynamic objects, our approach enables flexible scene manipulation, such as removing dynamic instances, showcasing its capability for high-fidelity reconstruction and editing of dynamic environments.
}
\label{fig:teaser}
\end{center}
}]

\renewcommand{\thefootnote}{\fnsymbol{footnote}}
\setcounter{footnote}{0}
\footnotetext[1]{Equal contribution to this work.}
\footnotetext[2]{Corresponding author (\url{lizhangfd@fudan.edu.cn}).}

\begin{abstract}
The realistic reconstruction of street scenes is critical for developing real-world simulators in autonomous driving. Most existing methods rely on object pose annotations, using these poses to reconstruct dynamic objects and move them during the rendering process. This dependence on high-precision object annotations limits large-scale and extensive scene reconstruction. To address this challenge, we propose \textbf{\bz{} curve Gaussian splatting} (\textbf{\model{}}),
which represents the motion trajectories of dynamic objects using learnable \bz{} curves. This approach fully leverages the temporal information of dynamic objects and, through learnable curve modeling, automatically corrects pose errors.
By introducing additional supervision on dynamic object rendering and inter-curve consistency constraints, we achieve reasonable and accurate separation and reconstruction of scene elements. Extensive experiments on the Waymo Open Dataset and the nuPlan benchmark demonstrate that \model{} outperforms state-of-the-art alternatives in both dynamic and static scene components reconstruction and novel view synthesis.
\end{abstract}   
\section{Introduction}
Modeling dynamic 3D street scenes is fundamental to modern autonomous driving, as it enables realistic and controllable simulations for tasks such as perception~\cite{liu2023bevfusion, chen2023voxelnext, yin2021centerpoint, li2023pillarnext}, prediction~\cite{prediction1, prediction2, prediction3}, and motion planning~\cite{plan1, plan2, plan3, RAD, xie2024vid2sim}. With the rise of end-to-end autonomous driving systems that require real-time sensor feedback~\cite{end-to-end1, end-tu-end2, end-to-end3}, the need for closed-loop evaluation of real-world simulations~\cite{open-loop-eval1,open-loop-eval2} has become more urgent. High-quality scene reconstruction creates the simulation environment for closed-loop assessment, while making it possible to safely and cost-effectively simulate critical extreme scenarios~\cite{ljungbergh2024neuroncap, zhou2024hugsim}.

Despite promising results in achieving realistic reconstruction of small-scale scenes, driving scenarios are large-scale and highly dynamic, making effective 3D scene modeling more challenging. To address these challenges, most existing methods~\cite{khan2024autosplat,wu2023mars,zhou2024drivinggaussian, nsg, yan2024street, zhou2024hugs, chen2024omnire, hess2024splatad} rely on manual pose annotations of dynamic objects to distinguish static backgrounds from moving objects. Typically, dynamic objects are reconstructed in their respective centered canonical spaces and then placed into the background scene space during rendering based on the known poses. However, manual annotations of dynamic objects are always subject to errors and omissions, limiting the applicability of these methods across diverse scenes in different datasets.

Other approaches that do not require dynamic annotations~\cite{huang2024s3gaussian, chen2023periodic} utilize self-supervised methods to learn the motion of dynamic objects. $S^3$Gaussian~\cite{huang2024s3gaussian} uses a spatial-temporal decomposition network to implicitly model the motion trajectories of objects, which introduces challenges in optimizing and modeling these trajectories. PVG~\cite{chen2023periodic} constructs long trajectories by stitching together segments with periodic vibration. However, the periodic vibration pattern and opacity decay do not align with the real-world motion, and segmenting trajectories makes it difficult to fully exploit the consistency of a single object over time.

To overcome the identified limitations, this paper introduces a novel dynamic scene representation method called \textbf{\bz{} curve Gaussian splatting (\model{})} to achieve high-fidelity novel view synthesis performance for autonomous driving applications. Based on efficient 3D Gaussian splatting techniques~\cite{kerbl20233d}, this method explicitly models the motion trajectories and velocity of dynamic Gaussian primitives in the scene using learnable \bz{} curves, while static 3D Gaussian primitives are used to construct the background information. The learnable trajectory curves can compensate for annotation errors in dynamic objects.
Furthermore, the explicit curve trajectories facilitate optimization and fully leverage the temporal consistency of the same object across different timestamps.
We group dynamic Gaussian primitives according to the reconstructed objects and introduce a grouped inter-curve consistency loss, effectively utilizing the geometric constraints of the same object. Additionally, we introduce extra supervision for the rendering of dynamic Gaussian primitives to enhance the reconstruction of the dynamic components, facilitating subsequent autonomous driving scene editing tasks.

Our main contributions are summarized as follows: \textbf{({\romannumeral 1})} We propose \bz{} curve Gaussian splatting (\model{}) for large-scale dynamic urban scene reconstruction.  Through explicit learnable \bz{} curve trajectory modeling, we elegantly represent dynamic scenes, eliminating the dependence on the precision of object annotations in street scene reconstruction.
\textbf{({\romannumeral 2})} We develop a novel grouped inter-curve consistency loss that links the trajectories of Gaussian primitives constituting the same object, effectively leveraging the geometric constraints of the same object.
\textbf{({\romannumeral 3})} Extensive experiments conducted on two large-scale benchmark datasets (Waymo~\cite{sun2020scalability} and nuPlan~\cite{karnchanachari2024towards}) demonstrate that \model{} outperforms all previous state-of-the-art alternatives in both scene reconstruction and novel view synthesis.
\begin{figure*}[t]
    \centering
    \includegraphics[width=\linewidth]{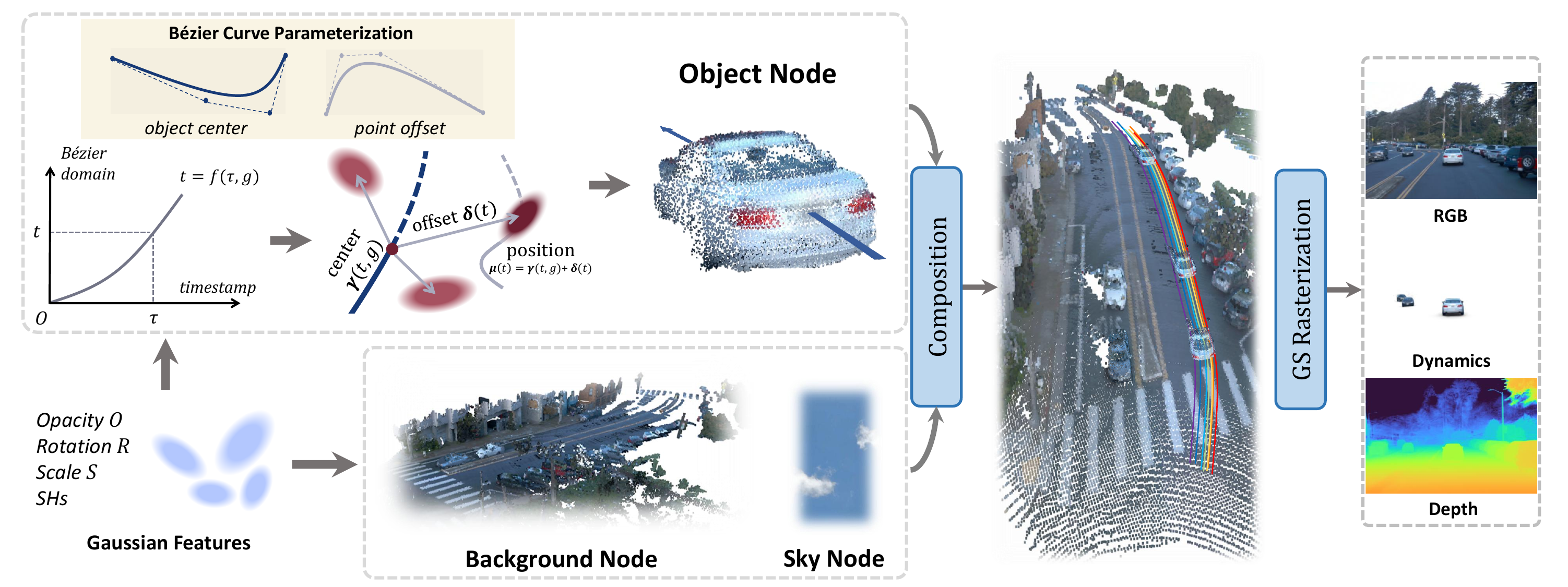}
    \vspace{-6.5mm}
    \caption{\textbf{Pipeline.} 
    The Gaussian trajectories of all foreground objects are controlled by learnable \bz{} curves.
    Due to the varying motion speeds of different dynamic objects, the time $\tau$ is mapped to different \bz{} parameters $t$ for each object $g$ to determine the positions of Gaussian primitives, which are then integrated with the static background Gaussian primitives and the sky model for rendering.
    We model the Gaussian trajectory as the sum of its corresponding object center $\vgamma(t,g) $ and an offset relative to the center $\vdelta(t)$, allowing us to exploit the trajectory consistency of the same object by constraining their offset trajectories.
    }
    \vspace{-5mm}
    \label{fig:pipeline}
\end{figure*}
\section{Related work}

\noindent{\bf NeRF for urban scene reconstruction.} 
Since the introduction of NeRF~\cite{mildenhall2020nerf}, neural representations have become a fundamental cornerstone in 3D reconstruction and novel view synthesis. Numerous works have applied NeRF-based methods to urban scene~\cite{nsg, turki2023suds, yang2023emernerf, nguyen2024rodus, wu2023mars, tonderski2024neurad, xie2023snerf, chen2025snerf, yang2023unisim}, enabling sensors to achieve realistic novel view rendering in large and dynamic scenes.  NSG~\cite{nsg} employs neural scene graphs to decompose dynamic scenes, while SUDS~\cite{turki2023suds} introduces a 4D scene representation using a multi-branch hash table. Self-supervised methods, such as EmerNeRF~\cite{yang2023emernerf} and RoDUS~\cite{nguyen2024rodus}, effectively address the challenges of dynamic scenes.  However, the slow rendering speed of NeRF-based methods presents significant challenges and high costs for their application in downstream tasks~\cite{ljungbergh2024neuroncap}. In contrast, \model{} reconstructs urban scenes using efficient 3D Gaussian primitives, achieving superior rendering quality while maintaining high rendering speed.

\noindent{\bf 3DGS for urban scene reconstruction.}
Recent studies have leveraged efficient 3D Gaussian splatting (3DGS)~\cite{kerbl20233d} techniques for urban scene reconstruction~\cite{chen2023periodic, huang2024s3gaussian, yan2024street, zhou2024drivinggaussian, zhou2024hugs, zhang2024street, chen2024omnire, peng2024desire, yang2024storm, jiang2025gslidar, zhou2024lidarrt}, achieving significant improvements in both reconstruction quality and rendering speed. 
$S^3$Gaussian~\cite{huang2024s3gaussian} and PVG~\cite{chen2023periodic} employ self-supervised learning methods to infer the trajectories of dynamic objects. Specifically, $S^3$Gaussian implicitly models the motion trajectories of objects using a spatio-temporal decomposition network, while PVG constructs long trajectories by concatenating segments exhibiting periodic vibration. However, these trajectory-modeling methods still lack precision.
Meanwhile, explicitly decomposing scenes into distinct entities has emerged as a prevalent practice, as used in works such as Street Gaussians~\cite{yan2024street}, DrivingGaussian~\cite{zhou2024drivinggaussian}, HUGS~\cite{zhou2024hugs}, and OmniRe~\cite{chen2024omnire}. However, these methods heavily rely on the accuracy of manually annotated bounding boxes, and their reconstruction performance significantly deteriorates when the annotations are imprecise. In contrast, \model{} explicitly and accurately represents the trajectories of dynamic objects using learnable \bz{} curves, thereby eliminating the dependence on the precision of object annotations in urban scene reconstruction.

\section{Method}

\begin{figure*}[t]
    \centering
    \includegraphics[width=0.9\linewidth]{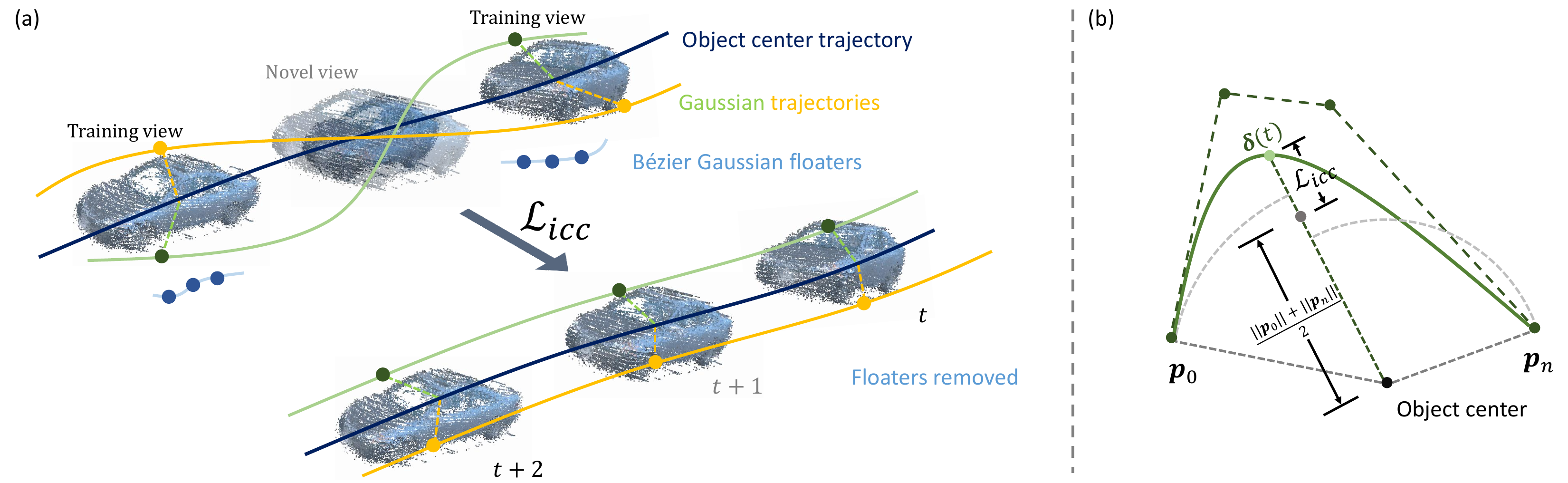}
    \vspace{-2mm}
    \caption{
    (a) Directly modeling Gaussian trajectories introduces excessive degrees of freedom, leading to uncontrolled motion and even drift into static elements, causing floaters and artifacts in novel view synthesis. To address this issue, we propose the inter-curve consistency loss $\mathcal{L}_{icc}$ to enforce trajectory consistency among Gaussian primitives of the same object, effectively eliminating floaters. (b) We constrain the offset $\vdelta(t)$ by using the average magnitude of its start $\vp_0=\vdelta(0)$ and end $\vp_n=\vdelta(1)$ points, reducing variations in the offset from the center, thereby ensuring the consistency between the Gaussian primitive's trajectory and the object's center trajectory.
    }
    \vspace{-15pt}
    \label{fig:group_offset}
\end{figure*}

In this section, we propose \model{}, a novel framework for achieving precise 3D reconstruction and synthesize novel viewpoints from any desired timestamp and camera pose in an urban scene.  
In~\Cref{preliminary}, we introduce the fundamental concepts of 3D Gaussian splatting and \bz{} curves. \Cref{beziergs} provides a detailed description of our proposed method, where we combine static and dynamic Gaussian primitives to reconstruct the background and foreground of the scene, respectively. The trajectories of dynamic Gaussian primitives are modeled using learnable \bz{} curves. Finally, in~\Cref{loss}, we discuss the various functions used to optimize the scenes, which enhance the geometric representation of objects and enable precise separation of dynamic and static components. An overview of our pipeline is provided in~\Cref{fig:pipeline}.

\subsection{Preliminaries}
\label{preliminary}
\noindent{\bf 3D Gaussian splatting.} 3D Gaussian splatting~\cite{kerbl20233d} (3DGS) utilizes a collection of 3D Gaussian primitives to represent a scene. Through a tile-based rasterization process, 3DGS facilitates real-time alpha blending of numerous Gaussian primitives. The scene is modeled by a set of Gaussian primitives, where each primitive contains the following attributes: mean position $\gaussmu \in \RR^3$, rotation $\gaussq \in \RR^4$ represented as a quaternion, anisotropic scaling factors $\gausss \in \RR^3$, opacity $\gausso \in [0, 1]$ and view-dependent colors $\gaussc \in \RR^{3}$ represented as spherical harmonics (SH) coefficients. 

To determine the pixel color $\tilde{\vc}\in\R^3$, the Gaussian primitives contributing to that pixel are first sorted based on their distance from the camera center (indexed by $i \in \mathcal{N}$) and then alpha blended:  
\begin{equation}
\tilde{\vc} = \sum_{i \in \mathcal{N}} {\gaussc}_{i} \alpha_{i} \prod_{j=1}^{i-1} (1 - \alpha_{j}).
\label{alpha blending}
\end{equation}  
Here, the opacity $\alpha_i$ is computed as:  
\begin{equation}
    \alpha_i = \gausso_i \exp\left(-\frac{1}{2} (\vxi - \bm{\mu}_i)^T \bm{\Sigma}_{i}^{-1} (\vxi - \bm{\mu}_i)\right),
\end{equation}
where $\vxi \in \mathbb{R}^2$ is the 2D pixel position in the image plane, $\vmu_i\in \R^2$ is the projected center of the $i$-th Gaussian and $ \mSigma_{i} \in \R^{2\times2}$ denotes the 2D projected covariance matrix.

\noindent{\bf \bz{} curves.} The \bz{} curve~\cite{mortenson1999mathematics} is a fundamental parametric curve in computer graphics and computational geometry, defined by $ n+1 $ control points $\{\vp_i\}_{i=0}^n$ and expressed as a function of the parameter $ t $ as follows:  
\begin{equation}
\vp(t)=\sum_{i=0}^{n} b_{i,n}(t) \vp_i, \quad t\in [0, 1],
\label{bezier motion}
\end{equation}  
where $ \vp(t) $ represents the position of the curve at a given parameter $ t $, $ \vp_i $ denotes the position of $ i $-th control point, and $ b_{i,n}(t) $ is the Bernstein basis polynomial of degree $ n $, which is defined as:  
\begin{equation}
b_{i,n}(t) = \binom{n}{i} t^i (1 - t)^{n-i}, \quad i \in \{0,1 \dots, n\}.
\end{equation}  
The parameter $ t $ varies within the interval $[0,1]$, where $ t=0 $ corresponds to the first control point $ \vp_0 $, and $ t=1 $ corresponds to the last control point $ \vp_n $. The curve is constructed as a weighted sum of the control points, with the Bernstein basis polynomials determining the influence of each control point at a given $ t $.

\subsection{\bz{} curve Gaussian splatting}
\label{beziergs}
In a 3D street scene, static and dynamic components exhibit different motion characteristics, necessitating the use of distinct Gaussian primitives for their representation. For the static background, since the background remains consistent across all frames, we can achieve a stable 3D representation through global optimization, unaffected by temporal changes. We use 3DGS~\cite{kerbl20233d} for reconstruction, where each Gaussian is characterized by the attributes $\{\gaussmu, \gaussq, \gausss, \gausso, \gaussc\}$, which is detailed in~\Cref{preliminary}.

For the dynamic foreground, we model the trajectory of Gaussian primitives using learnable \bz{} curves. Existing bounding box based methods~\cite{khan2024autosplat,wu2023mars,zhou2024drivinggaussian, nsg, yan2024street} heavily rely on the accuracy of the bounding boxes (orientation and position), which can be a limiting factor. 
For other methods that use self-supervised learning~\cite{huang2024s3gaussian, chen2023periodic} to model the motion trajectories of dynamic objects, it is difficult to ensure both the accuracy of the trajectories and training efficiency.
In contrast, our learnable \bz{} curves eliminate the dependence on the accuracy of manual annotations, while effectively representing the complete trajectories.

Since different objects in the scene follow distinct motion trajectories, we represent each object using a separate group of Gaussian primitives. Specifically, the trajectory of a Gaussian primitive is defined as the sum of its corresponding object's center and an offset relative to the center, where the offset is expressed in the world coordinate system. This representation enables us to control the trajectory consistency of different Gaussian primitives within the group by constraining the offsets.
To distinguish between distinct objects, we introduce an additional attribute, $g \in \mathbb{N_+}$, which characterizes the specific object that the Gaussian primitive reconstructs.
The trajectory of an object's center is modeled using a learnable \bz{} curve, controlled by a series of control points $\{\vp_i^g\}_{i=0}^n$. Given $t \in [0, 1]$, the center $\vgamma(t, g)\in\R^3$ of object $g$ is defined as:  
\begin{equation}
\vgamma(t, g) = \sum_{i=0}^{n} b_{i,n}(t) \vp_i^g \quad t \in [0, 1]
\end{equation}

To model the offset trajectory of the Gaussian primitives relative to the objects' center, we also use learnable \bz{} curves controlled by a set of control points. For a specific Gaussian primitive, the offset trajectory is defined by the control points $\{\vp_i\}_{i=0}^{n}$, where $\vp_i \in \mathbb{R}^3$ represents the position of the $i$-th control point. The offset $\vdelta(t)\in\R^3$ at $t$ is given by:  
\begin{equation}
    \vdelta(t) = \sum_{i=0}^{n} b_{i,n}(t) \vp_i
\end{equation}

The motion of objects along the \bz{} curves is non-uniform over time, making it necessary to model their velocity, which can be implicitly represented by the time-to-\bz{} mapping between timestamp $ \tau $ and \bz{} parameter $ t $. Additionally, for each object $ g $, the time-to-\bz{} mapping varies. To capture this variation, we consistently use additional \bz{} curves to model the time-to-\bz{} mapping $ t = f(\tau, g) $, as detailed in~\Cref{sec:t2b-mapping}.
In summary, the motion trajectory of the Gaussian primitive can be formulated as:
\begin{equation}
\vmu(\tau, g) = \vdelta(t) + \vgamma(t, g), \quad \text{where } t = f(\tau, g).   
\label{dynamic pos}
\end{equation}
Similarly, like the static part, our dynamic component also possesses the attributes $\{\gaussq, \gausss, \gausso, \gaussc\}$, which is detailed in~\Cref{preliminary}.

Given a recorded timestamp $\tau$, we compute the position of dynamic Gaussian primitives $\vmu(\tau, g)$ using \Cref{dynamic pos} and combine them with static Gaussian primitives. The final rendering of the RGB map $\mC_\gset$, depth map $\mD_\gset$, and opacity map $\mO_\gset$ is achieved using alpha-blending technique in \Cref{alpha blending}.

Since 3DGS~\cite{kerbl20233d} is defined in Euclidean space, it is not well-suited for modeling distant regions such as the sky. To address this, we use a high-resolution cube map that maps the view direction to the sky color $\mC_{\text{sky}}$ as the sky texture. By compositing the sky image $\mC_{\text{sky}}$ with the rendered Gaussian colors $\mC_{\gset}$, the final rendering is obtained as:  
\begin{equation}
\boldsymbol{C} = \boldsymbol{C}_{\mathcal{G}} + (\mathbf{1} - \boldsymbol{O}_{\mathcal{G}}) \odot \boldsymbol{C}_{\text{sky}}.
\end{equation}

We note that bounding box based methods~\cite{khan2024autosplat,wu2023mars,zhou2024drivinggaussian, nsg, yan2024street} represent a special case of \model{}, where the offset is defined in the object coordinate system and remains constant over time, while the bounding boxes' orientations and translations remain fixed.
Additionally, we can represent long trajectories using piecewise \bz{} curves, serving as a higher-level alternative to the periodic vibration characterization in PVG~\cite{chen2023periodic}. By extending the dynamic representation capability of Gaussian primitives, our model flexibly and accurately models trajectories (\Cref{fig:teaser}(a)), achieving state-of-the-art reconstruction performance.

\definecolor{best_result}{rgb}{0.96, 0.57, 0.58}
\definecolor{second_result}{rgb}{0.98, 0.78, 0.57}

\begin{table*}[tb]
    \centering
    \tiny
    \resizebox{\linewidth}{!}{
    \renewcommand{\arraystretch}{0.8}
    \begin{tabular}{l|l|cccc|cccc}
    \toprule
    \multirow{2}{*}{\textbf{Dataset}} & \multirow{2}{*}{\textbf{Methods}} & \multicolumn{4}{c|}{\textbf{Image Reconstruction}} & \multicolumn{4}{c}{\textbf{Novel View Synthesis}} \\
    & & PSNR$\uparrow$ & SSIM$\uparrow$ & LPIPS$\downarrow$ & Dyn-PSNR$\uparrow$ & PSNR$\uparrow$ & SSIM$\uparrow$ & LPIPS$\downarrow$ & Dyn-PSNR$\uparrow$ \\
    \midrule
    \multirow{6}{*}{Waymo~\cite{sun2020scalability}} 
     & DeformableGS~\cite{yang2024deformable} & 32.34 & \cellcolor{second_result}0.923 & 0.086 & 27.39 & 29.52 & \cellcolor{second_result}0.889 & \cellcolor{second_result}0.100 & 24.66 \\
     & HUGS~\cite{zhou2024hugs}  & 31.58 & 0.904 & 0.094 & 25.78 & 29.34 & 0.865 & 0.110 & 23.84 \\
     & Street Gaussians~\cite{yan2024street} & 32.18 & 0.918 & 0.090 & 28.64 & 28.92 & 0.877 & 0.110 & 25.54 \\
     & OmniRe~\cite{chen2024omnire} & 32.32 & \cellcolor{second_result}0.923 & \cellcolor{second_result}0.084 & 28.36 & 29.41 & 0.884 & 0.101 & \cellcolor{second_result}25.85 \\
     & PVG~\cite{chen2023periodic} & \cellcolor{second_result}32.61 & 0.901 & 0.159 & \cellcolor{second_result}29.32 & \cellcolor{second_result}29.64 & 0.864 & 0.179 & 24.46 \\
     & \textbf{\model{} (Ours)} & 
     \cellcolor{best_result}\textbf{33.98} & 
     \cellcolor{best_result}\textbf{0.934} & 
     \cellcolor{best_result}\textbf{0.077} & 
     \cellcolor{best_result}\textbf{32.39} & 
     \cellcolor{best_result}\textbf{31.51} & 
     \cellcolor{best_result}\textbf{0.903} & 
     \cellcolor{best_result}\textbf{0.092} & 
     \cellcolor{best_result}\textbf{28.51} \\
    \midrule
    \multirow{6}{*}{nuPlan~\cite{karnchanachari2024towards}} 
     & DeformableGS~\cite{yang2024deformable} & 28.10 & \cellcolor{second_result}0.880 & \cellcolor{second_result}0.143 & 22.64 & 26.20 & \cellcolor{second_result}0.824 & \cellcolor{second_result}0.159 & 21.37 \\
     & HUGS~\cite{zhou2024hugs} & 24.16 & 0.768 & 0.274 & 20.39 & 23.77 & 0.744 & 0.280 & 20.15 \\
     & Street Gaussians~\cite{yan2024street} &   26.05 & 0.813 & 0.180 & 22.19 & 25.76 & 0.788 & 0.185 & 22.03 \\
     & OmniRe~\cite{chen2024omnire} & 26.72 & 0.845 & 0.165 & \cellcolor{second_result}24.28 & 26.01 & 0.819 & 0.173 & \cellcolor{second_result}23.90 \\
     & PVG~\cite{chen2023periodic} & \cellcolor{second_result}29.08 & 0.869 & 0.170 & 21.94 & \cellcolor{second_result}26.38 & 0.772 & 0.222 & 19.69 \\
     & \textbf{\model{} (Ours)} & 
     \cellcolor{best_result}\textbf{30.74} & 
     \cellcolor{best_result}\textbf{0.896} & 
     \cellcolor{best_result}\textbf{0.122} & 
     \cellcolor{best_result}\textbf{27.13} & 
     \cellcolor{best_result}\textbf{29.42} & 
     \cellcolor{best_result}\textbf{0.860} & 
     \cellcolor{best_result}\textbf{0.133} & 
     \cellcolor{best_result}\textbf{25.12} \\
    \bottomrule
    \end{tabular}
    }
    \vspace{-3mm}
    \caption{\textbf{Comparison to state-of-the-art methods on Waymo~\cite{sun2020scalability} and nuPlan~\cite{karnchanachari2024towards}.} We color the top results as \colorbox{best_result}{best} and \colorbox{second_result}{second best}.}
    \vspace{-15pt}
    \label{tab:waymo_nuplan_metrics}
\end{table*}
\subsection{Loss functions}
\label{loss}
The overall loss is formulated as:
\begin{align}
\mathcal{L} = & (1-\lambda_{r})\mathcal{L}_1 + \lambda_r \mathcal{L}_{\mathrm{ssim}} + \lambda_{o}^{sky} \mathcal{L}_{o}^{sky} + \nonumber \\  
        & \lambda_{icc}\mathcal{L}_{icc} + \lambda_{dr}\mathcal{L}_{dr} + \lambda_{v}\mathcal{L}_{v} + \lambda_d \mathcal{L}_d
\end{align}
where $\mathcal{L}_1$ and $\mathcal{L}_{\text{ssim}}$ represent the L1 and SSIM losses, respectively, which are used to supervise the RGB rendering quality. The depth loss $\mathcal{L}_d$ is introduced to enhance geometry awareness and is defined as: $\mathcal{L}_d = \left\|\mD - \mD_{\gset}\right\|_1$, where $\mD$ is a sparse inverse depth map obtained by projecting LiDAR points onto the camera plane, and $\mD_{\gset}$ denotes the inverse of the rendered depth map. Additionally, we introduce $\mathcal{L}_{O}^{sky}$ to reduce opacity in the sky region: $\mathcal{L}_{O}^{sky} = -\sum \mM^{sky} \cdot \log (1 - \mO_{\gset})$, where $\mM^{sky}$ is a binary sky mask predicted by the Grounded-SAM~\cite{ren2024grounded} model. This term encourages the rendered opacity map $\mO_{\gset}$ to be minimized in the sky region, ensuring that the sky is modeled only using the sky texture. 

\noindent{\bf Inter-curve consistency loss.} 
During the optimization process, due to the high degrees of freedom of Gaussian primitives, a single primitive may uncontrollably deviate from the dynamic object it represents.
This results in certain regions of the object being represented by different primitives at different time steps, leading to inconsistencies when rendered from novel viewpoints, as shown in~\Cref{fig:group_offset}.
To address this issue, it is crucial to enhance the temporal geometric consistency of Gaussian primitives.

For dynamic objects in a scene, as they move as a whole, the difference between the trajectory of a specific part and that of the entire object remains within a limited range.  Specifically, for rigid structures such as vehicles, the magnitude of trajectory deviation tends to remain constant.  Therefore, by ensuring consistency in the offset trajectory $\vdelta(t)$ of Gaussian primitive over time, the similarity between the Gaussian primitive's trajectory $\vmu(\tau, g) = \vdelta(t) + \vgamma(t, g)$ and the corresponding object center trajectory $\vgamma(t, g)$ can be maintained, thereby preserving the temporal coherence of the dynamic Gaussian primitive's representation of object parts. Since Gaussian primitives coincide with the first and last control points $\{\vp_0,\vp_n\}$ when the \bz{} curve parameter $t$ is set to 0 or 1, the offset $\vdelta(t)$ at a given timestamp $\tau$ can be constrained by the average magnitude of the first and last control points of the offset curve:
\begin{equation}
\mathcal{L}_{icc}=\left\|\|\vdelta(t)\|-\frac{\|\vp_0\|+\|\vp_n\|}{2} \right\|_1  
\end{equation}
This loss function effectively suppresses excessive local geometric variations, enhancing the overall structural consistency and stability.

\noindent{\bf Dynamic rendering loss.} 
Due to the alpha-blending mechanism described in~\Cref{alpha blending}, interactions between dynamic  and static Gaussian primitives may introduce mutual interference, making it challenging to accurately model dynamic and static elements in the scene separately. To address this issue, we introduce additional supervision on the rendering results of dynamic Gaussian primitives, ensuring that the rendering of dynamic scene components is solely contributed by the dynamic Gaussian primitives.

To obtain accurate masks for dynamic objects in the scene, we first project manual or automatic labeled dynamic 3D bounding boxes into images to extract the dynamic regions, and then obtain accurate masks for dynamics in each region by Grounded-SAM~\cite{ren2024grounded}, which are annotated as $\mM^{dyn}$. We use $\mM^{dyn}$ to extract the dynamic components of ground-truth camera images, which are then employed to supervise the rendered RGB map of dynamic Gaussian primitives:
\begin{equation}
    \mathcal{L}_{rgb}^{dyn}=\left(1-\lambda_r\right) \mathcal{L}_1^{dyn}+\lambda_r \mathcal{L}_{\mathrm{ssim}}^{dyn}.
\end{equation}
where  $\mathcal{L}_1^{dyn}$ and $\mathcal{L}_{\text{ssim}}^{dyn}$ represent the L1 and SSIM losses for masked ground-truth camera image and the rendered RGB map of dynamic Gaussian primitives.

To further enhance the separation between dynamic and static scene components, we introduce an additional constraint to ensure that the rendered alpha map of the dynamic Gaussian primitives $\mO^{dyn}_{\gset}$ aligns with the mask of the dynamic part in the camera image $\mM^{dyn}$:
\begin{equation}
    \mathcal{L}_o^{dyn} = \left\| \mM^{dyn} - \mO^{dyn}_{\gset} \right\|_1
\end{equation}

By combining these two loss functions, we obtain the dynamic rendering loss:
\begin{equation}
\mathcal{L}_{dr}=\mathcal{L}_{rgb}^{dyn}+\mathcal{L}_o^{dyn}
\end{equation}
This loss function ensures that the rendering of dynamic scene components is solely contributed by the dynamic Gaussian primitives, leading to a more thorough separation between dynamic and static components of the scene, resulting in improved rendering quality when synthesizing from new viewpoints.

\noindent{\bf Velocity loss.}
To impose multi-dimensional constraints on the reconstruction of dynamic components, we add an additional constraint on the velocity map $\mV_{\gset}^{dyn}$ rendered by dynamic Gaussian primitives, ensuring that the motion trends of the Gaussian primitives align with those of the dynamic objects, thereby enhancing the plausibility of the dynamic representation.

From~\Cref{bezier motion}, the derivative of the position $\vp(t)$ on the \bz{} curve with respect to the parameter $t$ is given by:
\begin{gather}
    \frac{\mathrm{d} \vp(t)}{\mathrm{d}t} = \sum_{i=0}^n  \frac{\mathrm{d} b_{i,n}(t)}{\mathrm{d}t} \vp_i, \\
    \text{where}\ \frac{\mathrm{d} b_{i,n}(t)}{\mathrm{d}t} = \binom{n}{i}(i-nt) t^{i-1}(1-t)^{n-i-1}.
\end{gather}
As both the object center trajectory $\vgamma(t, g)$ and the offset $\vdelta(t)$ are modeled using \bz{} curves, the velocity $\vv(\tau,g)$ of the Gaussian primitive at a given timestamp $\tau$ is:
\begin{equation}
\vv(\tau,g) = \left( \frac{\mathrm{d} \vgamma(t,g)}{\mathrm{d}t} + \frac{\mathrm{d} \vdelta(t)}{\mathrm{d}t} \right) \frac{\mathrm{d}t}{\mathrm{d}\tau}
\end{equation}
which subsequently used to render the velocity map $V_{\gset}^{dyn}$ of dynamic Gaussian primitives as follows:
\begin{equation}
    \mV_\gset^{dyn} = \sum_{i \in \mathcal{N}} \vv_i(\tau, g) \alpha_i \prod_{j=1}^{i-1} \left( 1 - \alpha_j \right)
\end{equation}
To ensure that the motion of dynamic Gaussian primitives is strictly confined within the dynamic region $M^{dyn}$, we introduce the following loss function:
\begin{equation}
    \mathcal{L}_{v}=\left\| \mV_{\gset}^{dyn} \cdot(1-\mM^{dyn}) \right\|_2
\end{equation}
By reasonably controlling the velocity of the dynamic Gaussian primitives, we implicitly prevent the dynamic Gaussian primitives from drifting into static regions, ensuring that their movement is confined within the dynamic objects, thereby further enhancing the reliability of the dynamic representation.

\section{Experiments}
\begin{figure*}[t]
    \centering
    \includegraphics[width=\linewidth]{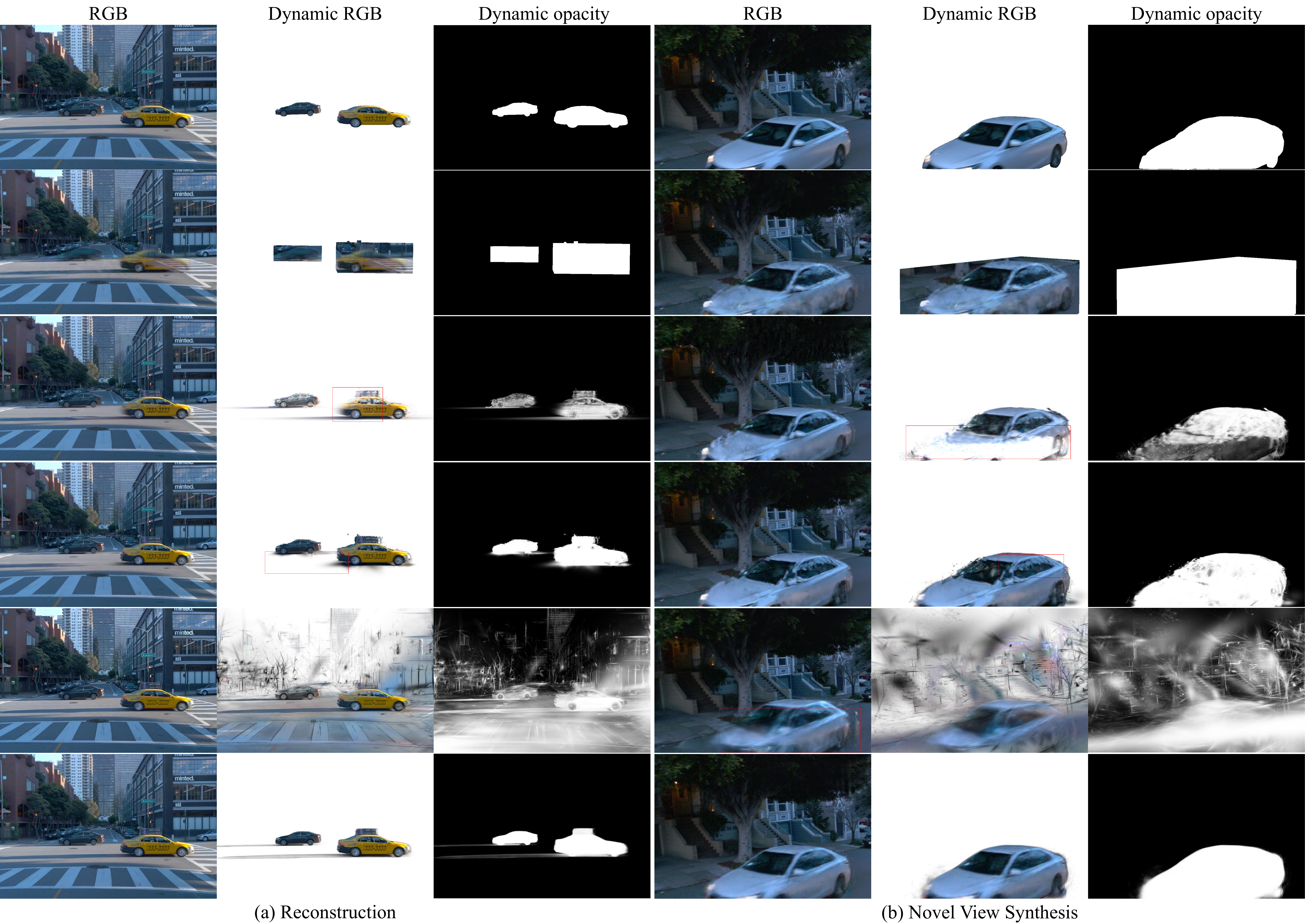}

    \vspace{-120mm}
    \caption*{
        \hspace{-178mm}\rotatebox{90}{\fontsize{6}{7.2}\selectfont Ground truth}}
    \vspace{1mm}
    \caption*{
        \hspace{-178mm}\rotatebox{90}{\fontsize{6}{7.2}\selectfont DeformableGS~\cite{yang2024deformable}}
    }
     \vspace{-2mm}
    \caption*{
        \hspace{-178mm}\rotatebox{90}{\fontsize{6}{7.2}\selectfont Street Gaussians~\cite{yan2024street}}
    }
     \vspace{1mm}
    \caption*{
        \hspace{-178mm}\rotatebox{90}{\fontsize{6}{7.2}\selectfont OmniRe~\cite{chen2024omnire}}
    }
     \vspace{5mm}
    \caption*{
        \hspace{-178mm}\rotatebox{90}{\fontsize{6}{7.2}\selectfont PVG~\cite{chen2023periodic}}
    }
     \vspace{5mm}
    \caption*{
        \hspace{-178mm}\rotatebox{90}{\fontsize{6}{7.2}\selectfont \model{} (Ours)}
    }
    \vspace{2mm}
    \caption{\textbf{Waymo dataset Qualitative Comparison.} \model{} 
    effectively reconstructs both static and dynamic elements with high fidelity, while achieving a clean separation between static and dynamic components.}
    \label{fig:waymo_compare}
    \vspace{-4mm}
\end{figure*}

\subsection{Experimental Setup}
\noindent{\bf Datasets.}
We conduct experiments on the Waymo Open Dataset~\cite{sun2020scalability} and the nuPlan benchmark~\cite{karnchanachari2024towards}, both of which have a frame rate of 10 Hz. Notably, nuPlan is the world's first large-scale planning benchmark for autonomous driving. However, due to inaccuracies in its manual annotations, traditional bounding box based methods~\cite{yan2024street, chen2024omnire} struggle to achieve high-quality reconstruction, limiting the development of reconstruction-based closed-loop simulation. By demonstrating high-quality reconstruction and novel view synthesis on nuPlan, we further validate that our method eliminates the dependency on accurate object annotations in urban scene reconstruction.
For Waymo, we select 12 sequences chosen by Street Gaussians~\cite{yan2024street} and PVG~\cite{chen2023periodic}. For nuPlan, we select 6 sequences partitioned by NAVSIM~\cite{Dauner2024NEURIPS}. Consistent with Street Gaussians~\cite{yan2024street}  and OmniRe~\cite{chen2024omnire}, we use every 4th image in a sequence as test frames, while the remaining images are used for training.

\noindent{\bf Baseline methods.}
We evaluated our method against state-of-the-art approaches, including both bounding box-based methods HUGS~\cite{zhou2024hugs}, Street Gaussians~\cite{yan2024street}, OmniRe~\cite{chen2024omnire} and dynamic Gaussian primitive-based methods DeformableGS~\cite{yang2024deformable}, PVG~\cite{chen2023periodic}.

\noindent{\bf Implementation details.}
In this work, we focus on the standard cubic \bz{} curve ($n=3$), which has been widely recognized for its effectiveness in trajectory modeling~\cite{feng2022rethinking}. All experiments run on a single NVIDIA RTX A6000 for 30,000 iterations. 
We maintain a learning rate similar to the original 3DGS implementation while setting the regularization coefficients as: $\lambda_r=0.2$, $\lambda_d=1.0$, $\lambda_{o}^{sky}=0.05$, $\lambda_{icc}=0.01$, $\lambda_{dr}=0.1$, $\lambda_{v}=1.0$. For more details on the implementation, please refer to~\Cref{sec:supp-detial}.

\begin{figure*}[h]
    \centering
    \includegraphics[width=\linewidth]{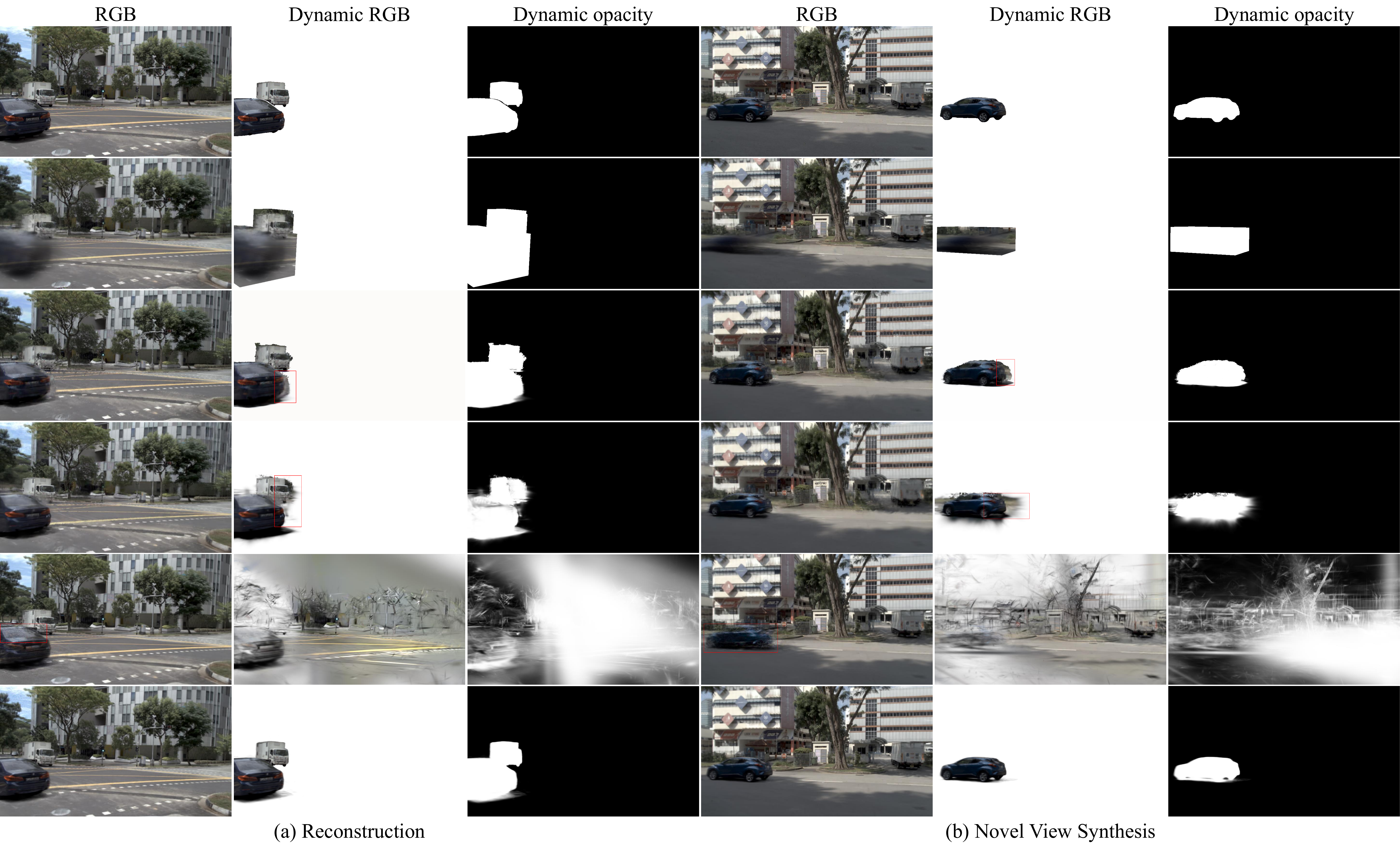}
    \vspace{-106mm}
    \caption*{
        \hspace{-178mm}\rotatebox{90}{\tiny Ground truth}}
    \vspace{-0.5mm}
    \caption*{
        \hspace{-178mm}\rotatebox{90}{\tiny DeformableGS~\cite{yang2024deformable}}
    }
     \vspace{-2.5mm}
    \caption*{
        \hspace{-178mm}\rotatebox{90}{\tiny 
        Street Gaussians~\cite{yan2024street}}
    }
     \vspace{0mm}
    \caption*{
        \hspace{-178mm}\rotatebox{90}{\tiny
        OmniRe~\cite{chen2024omnire}}
    }
     \vspace{4.5mm}
    \caption*{
        \hspace{-178mm}\rotatebox{90}{\tiny
        PVG~\cite{chen2023periodic}}
    }
    \vspace{3mm}
    \caption*{
        \hspace{-178mm}\rotatebox{90}{\tiny
        \model{} (Ours)}
    }
    \vspace{2mm}
    \caption{\textbf{nuPlan dataset Qualitative Comparison.} \model{} 
    can automatically correct pose errors of bounding boxes, resulting in improved reconstruction and novel view synthesis quality.
    }
    \vspace{-5.5mm}
    \label{fig:nuplan_compare}
\end{figure*}

\subsection{Comparisons with the State-of-the-art}

\noindent{\bf Results on Waymo.}
In addition to the standard PSNR, SSIM, and LPIPS metrics, we further assess the fidelity of dynamic regions by computing Dyn-PSNR, which specifically measures PSNR within the ground truth 3D bounding boxes projected onto the 2D image plane.
This additional evaluation provides a more precise assessment of reconstructing the dynamic elements. 
As shown in~\Cref{tab:waymo_nuplan_metrics}, \model{} outperforms state-of-the-art approaches~\cite{chen2023periodic, yan2024street, chen2024omnire, yang2024deformable, zhou2024hugs} across all evaluated metrics. Specifically, 
in novel view synthesis, our model effectively generates high-quality views at previously unseen timestamps, significantly surpassing all competing methods with a substantial 1.87~dB increase in PSNR, a 0.014 improvement in SSIM, and an 8.00\% reduction in LPIPS. Further, our approach outperforms alternatives in Dyn-PSNR, achieving a remarkable 2.66~dB gain, further validating its effectiveness in rendering dynamic content.

We present qualitative comparison with DeformableGS~\cite{yang2024deformable}, Street Gaussians~\cite{yan2024street}, OmniRe~\cite{chen2024omnire} and PVG~\cite{chen2023periodic} in~\Cref{fig:waymo_compare}.
Notably, DeformableGS~\cite{yang2024deformable} cannot separate dynamic objects. To address this, we extract dynamic regions by utilizing annotated bounding boxes.
As shown in~\Cref{fig:waymo_compare}, DeformableGS~\cite{yang2024deformable} struggles to effectively represent dynamic objects, while PVG~\cite{chen2023periodic} fails to effectively separate static and dynamic components.
Street Gaussians~\cite{yan2024street} and OmniRe~\cite{chen2024omnire} exhibit limitations in capturing dynamic objects, particularly leading to noticeable blurring around dynamic elements.
In contrast, our method effectively reconstructs both static and dynamic elements with high fidelity, while achieving a clean separation between static and dynamic components. 

\noindent{\bf Results on nuPlan.}
The nuPlan benchmark~\cite{karnchanachari2024towards} provides suboptimal bounding boxes, leading to a significant drop in rendering quality for bounding box based methods~\cite{zhou2024hugs, yan2024street, chen2024omnire}. In contrast, our \model{} models dynamic elements using learnable \bz{} curves, which enable automatically pose errors correction, resulting in improved reconstruction and novel view synthesis quality.
In novel view synthesis, our model effectively captures the scene and produces high-fidelity renderings, achieving a 3.04~dB improvement in PSNR, a 0.036 increase in SSIM, a 16.35\% reduction in LPIPS, and a 1.22~dB enhancement in Dyn-PSNR. The qualitative comparisons in \Cref{fig:nuplan_compare} further illustrate the effectiveness of our approach in handling complex dynamic scenes within the nuPlan benchmark~\cite{karnchanachari2024towards}.

\begin{table}[h]
\vspace{-1.5mm}
\centering
\footnotesize
\renewcommand{\arraystretch}{0.78}
\resizebox{\linewidth}{!}{
\begin{tabular}{clcccc@{}} 
\toprule
& & PSNR$\uparrow$  & SSIM$\uparrow$  & LPIPS$\downarrow$ & Dyn-PSNR $\uparrow$  \\
\midrule
(a) & w/o $\mathcal{L}_{\text{icc}}$
& 30.83 & 0.900 & 0.096 & 26.15 \\
(b) & w/o $\mathcal{L}_{\text{dr}}$ 
& 30.99 & 0.891 & 0.099 & 28.07 \\
(c) & w/o $\mathcal{L}_{\text{v}}$
& 31.40 & 0.901 & 0.094 & 28.29 \\
(d) & w/o time-to-\bz{}
& 31.36 & 0.899 & 0.094 & 27.97 \\
(e) & w/ MLP trajectory (DeformableGS)
& 29.58 & 0.898 & 0.087 & 24.78 \\
(f) & w/ sinusoidal trajectory (PVG)
& 29.65 & 0.877 & 0.099 & 26.27 \\ 
\midrule
& \textbf{\model{} (Ours)}
& \textbf{31.51} & \textbf{0.903} & \textbf{0.092} & \textbf{28.51} \\
\bottomrule
\end{tabular}}
\vspace{-3mm}
\caption{\textbf{Ablation Study} for novel view synthesis on Waymo.}
\vspace{-4mm}
\label{tab:ablation tab}
\end{table}

\subsection{Ablation study}
In~\Cref{tab:ablation tab}, we validate the effectiveness of key components of our method by measuring their impact on novel view synthesis metrics on Waymo~\cite{sun2020scalability}.
We see that \textbf{(a)} the inter-curve consistency loss enhances the ability to model dynamic objects and eliminates floaters, significantly improving  performance in novel view synthesis; \textbf{(b)} the dynamic rendering loss encourages dynamic objects to be modeled exclusively by dynamic Gaussian primitives, leading to a more thorough foreground-background separation; \textbf{(c)} the velocity loss further constrains the drift of dynamic Gaussian primitives, preventing interference with static Gaussian primitives; and \textbf{(d)} although the time-to-\bz{}  mapping provides a relatively smaller improvement, it is essential for modeling objects with highly complex trajectories in the scene.

\noindent{\bf Effectiveness of \bz{}.}
We replaced the dynamic trajectory modeling with MLP (DeformableGS~\cite{yang2024deformable}) or sinusoidal trajectory (PVG~\cite{chen2023periodic}), while maintaining the background reconstrunction with 3DGS~\cite{kerbl20233d} and all the losses except the inter-curve consistency loss. As shown in \Cref{fig:traj} and \Cref{tab:ablation tab}(e)(f), \bz{} curves enable more explicit and reasonable trajectory representation.

\section{Conclusion}
We have proposed the \bz{} curve Gaussian splatting (\model{}), an explicit scene representation for dynamic urban street scene reconstruction. By employing explicitly learnable \bz{} curves to model the motion trajectories of dynamic objects, our model can automatically correct pose errors, thus eliminating the reliance on the accuracy of manual annotations. The introduction of inter-curve consistency constraints enhances the temporal and geometric consistency of the dynamic Gaussian primitives. With additional supervision on the rendering of dynamic objects, our method enables the rational and accurate separation and reconstruction of scene elements. Our approach significantly outperforms the state-of-the-art methods on both the Waymo Open Dataset and the nuPlan benchmark.

\section*{Acknowledgments}
This work was supported in part by National Natural Science Foundation of China (Grant No. 62376060).
{
    \small
    \bibliographystyle{ieeenat_fullname}
    \bibliography{main}
}
\appendix
\begin{figure*}[h]
    \centering
    \includegraphics[width=\linewidth]{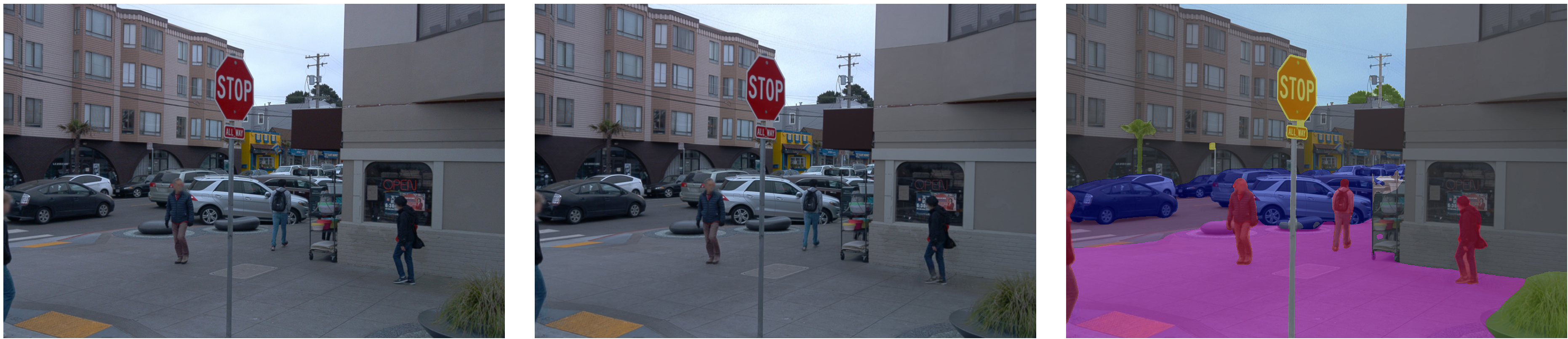}
    \vspace{-7mm}
    \caption*{
    \begin{tabularx}{0.95\linewidth}{>{\centering\arraybackslash}p{0.25\linewidth}>{\centering\arraybackslash}p{0.38\linewidth}>{\centering\arraybackslash}X}
       (a) Ground truth &  
       (b) Render &
       (c) Segmentation
    \end{tabularx}
}
    \vspace{-3mm}
    \caption{\textbf{Reconstruction of complex motions.} Although the reconstructed pedestrian legs of \model{} exhibit artifacts, the accurate semantic segmentation validates the correctness of visual perception for closed-loop simulation applications.
    }
    \vspace{-3mm}
    \label{fig:ped_recon}
\end{figure*}
\section*{Appendix}
\section{Implementation Details}
\label{sec:supp-detial}
\subsection{Initialization}
\subsubsection{\bz{} curve fitting}
For points $\{\vx_i\in\R^d\}_{i=0}^m$ that need to be initialized as a \bz{} curve of degree $n$, we first reasonably initialize the \bz{} parameters $\{t_i^0\}_{i=0}^m$ and solve for the control points' coordinates $\{\vp_i\in\R^d\}_{i=0}^n$ using the least squares method. Then, we compute the closest \bz{} parameter $t_i^1$ for each point $\vx_i$ based on the obtained \bz{} curve. This process is iteratively optimized until convergence. The workflow of our algorithm is presented in~\Cref{alg:bz_init}.

\begin{algorithm}[!h]
    \caption{\bz{} initialization}
    \label{alg:bz_init}
    \renewcommand{\algorithmicrequire}{\textbf{Input:}}
    \renewcommand{\algorithmicensure}{\textbf{Output:}}
    \begin{algorithmic}[1]
        \REQUIRE points $\{\vx_i\}_{i=0}^m$ to be initialized as a \bz{} curve
        \ENSURE control points $\{\vp_i\}_{i=0}^n$ of \bz{} curve 
        
        \STATE  reasonably initialize the \bz{} parameters $\{t_i^0\}_{i=0}^m$
        \STATE $k=0$
        \WHILE{not converge}
            \STATE solve $\{\vp_i^k\}_{i=0}^n$ by $\{\vx_i\}_{i=0}^m$ and $\{t_i^k\}_{i=0}^m$
            \STATE get  $\{t_i^{k+1}\}_{i=0}^m$
            \STATE $k=k+1$
        \ENDWHILE
        
        \RETURN $\{\vp_i^k\}_{i=0}^n$
    \end{algorithmic}
\end{algorithm}

\noindent{\textbf{\bz{} parameters initialization.}}
We leverages chord length parameterization to initialize the initial \bz{} parameter $\{t_i^0\}_{i=0}^m$. Given points $\{\vx_i\}_{i=0}^m$, we first compute the distances between consecutive points:
\begin{equation}
d_i = \|\vx_{i+1} - \vx_i\|, \quad i \in \{0,1 \dots, m-1\}.
\end{equation}
The parameter $\{t_i^0\}_{i=0}^m$ are then assigned based on the cumulative chord length:
\begin{equation}
t_0^0=0,\ t_i^0 = \frac{\sum_{k=0}^{i-1} d_k}{\sum_{k=0}^{n-1} d_k}, \quad i \in \{1,2 \dots, n\}.
\end{equation}

\noindent{\textbf{Control points fitting.}} 
Given the \bz{} parameters $\{t_i^k\}_{i=0}^m$ and corresponding points $\{\vx_i\}_{i=0}^m$, the control points $\{\vp_i^k\}_{i=0}^n$ can be solved by minimizing the distance between the curve and the points. This can be formulated as a least squares problem:
\begin{gather}
\min ||\mB\mP^T-\mX^T||
\end{gather}
where $\mB_{ij}=b_{i,n}(t_j^k)$, $\mP=\{\vp_0^k,\dots,\vp_n^k\}$ and $\mX=\{\vx_0^k,\dots,\vx_m^k\}$. $ b_{i,n}(t) $ is the Bernstein basis polynomial of degree $ n $, which is defined as:  
\begin{equation}
b_{i,n}(t) = \binom{n}{i} t^i (1 - t)^{n-i}, \quad i \in \{0,1 \dots, n\}.
\end{equation}  
Suppose that the matrix $\mB$ has full column rank, then the solution for $\mP$ can be expressed as $\mP=\mX\mB(\mB^T\mB)^{-1}$.

\subsubsection{Trajectories initialization}
\label{sec:traj_init}
For sequences with manual annotations, we use the annotated 3D bounding boxes to extract and merge LiDAR points from multiple frames of the same object, forming the initial point cloud of the object. The trajectory points $\{\vx_i\}$ are obtained by transforming the merged LiDAR points into the world coordinate system, followed by~\Cref{alg:bz_init} to get the corresponding control points $\{\vp_i\}$. 
For sequences without annotations, we employ a video instance tracking model~\cite{wu2024general} to extract the LiDAR points projected into the instance mask for each frame, and merge them as the initial point cloud of this instance. The center is computed by averaging the LiDAR point coordinates corresponding to each frame. Since our Bezier curve modeling can automatically correct errors, even this rough initialization leads to satisfactory reconstruction results.

\subsubsection{Time-to-\bz{} initialization}
\label{sec:t2b-mapping}
In the fitting process of~\Cref{sec:traj_init}, we simultaneously obtain the \bz{} parameter $\{t_i\}$ corresponding to each point. By combining $\{t_i\}$ with the timestamp $\{\tau_i\}$, we obtain a new set of points $\{\vx_i=(t_i, \tau_i)^T\}$ that need to be fitted. Similarly, we use~\Cref{alg:bz_init} to fit a \bz{} curve and thus model the time-to-\bz{} mapping.

\begin{figure*}[t]
\centering
\includegraphics[width=\linewidth]{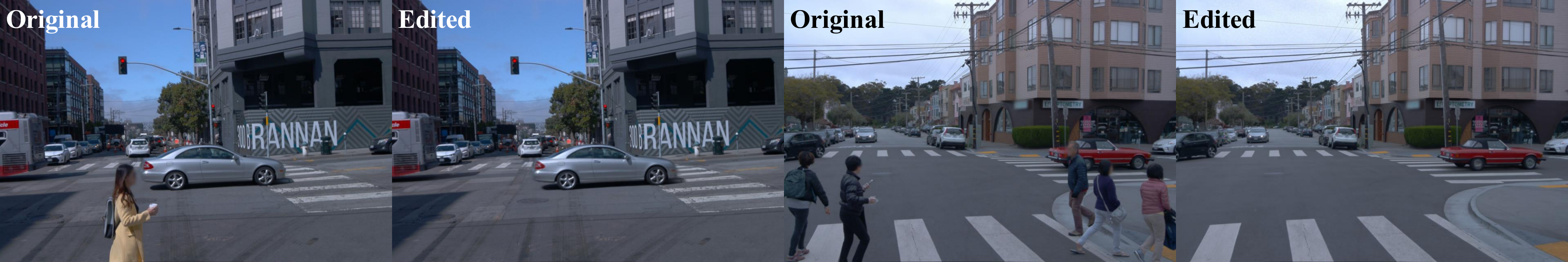}
\vspace{-7mm}
\caption{
Remove the pedestrians.
}
\label{fig:remove_ped}
\vspace{-1mm}
\end{figure*}
\begin{figure*}[t]
    \begin{minipage}[t]{0.6\linewidth}
        \centering
        \includegraphics[height=0.205\textheight]{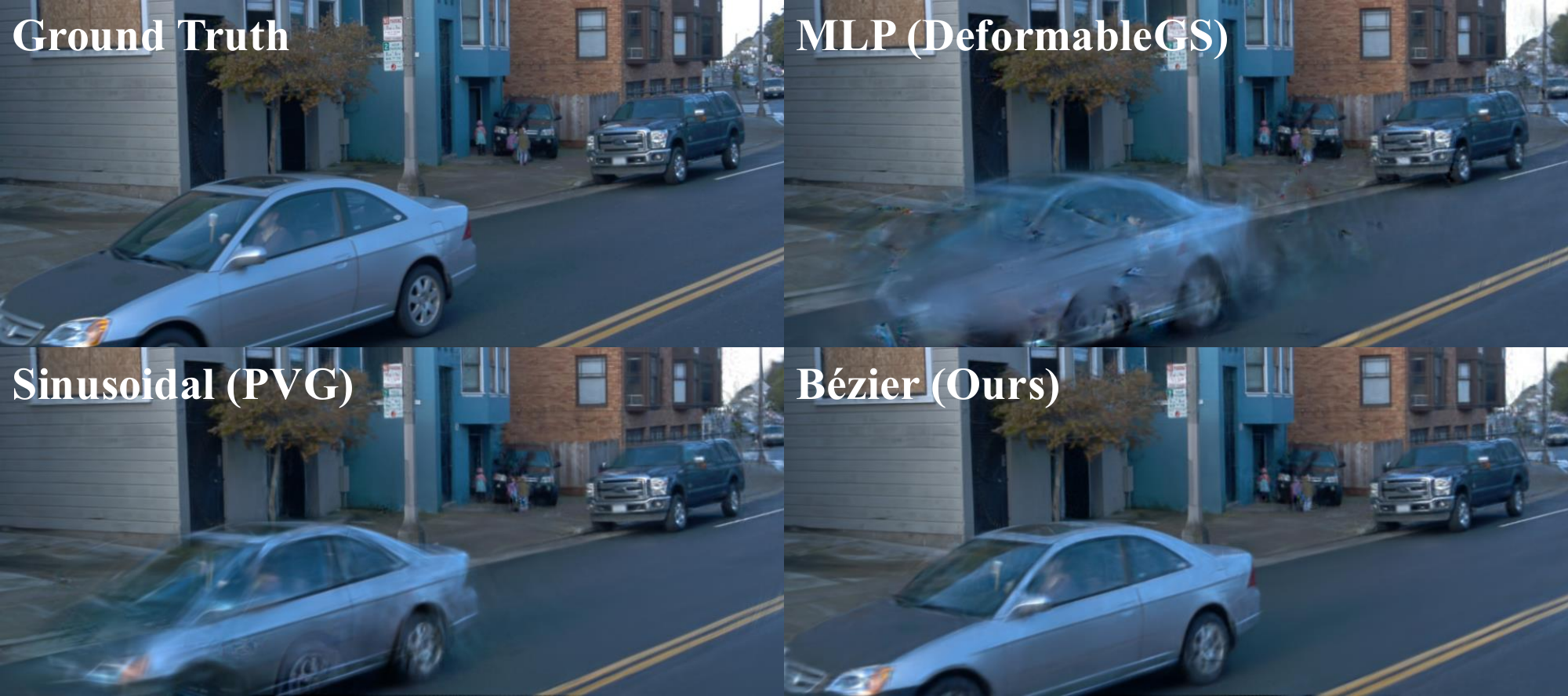}
        \vspace{-6mm}
        \caption{\textbf{Dynamic trajectory ablation} in novel view synthesis.}
        \label{fig:traj}
    \end{minipage}%
    \begin{minipage}[t]{0.4\linewidth}
        \centering
        \includegraphics[height=0.205\textheight]{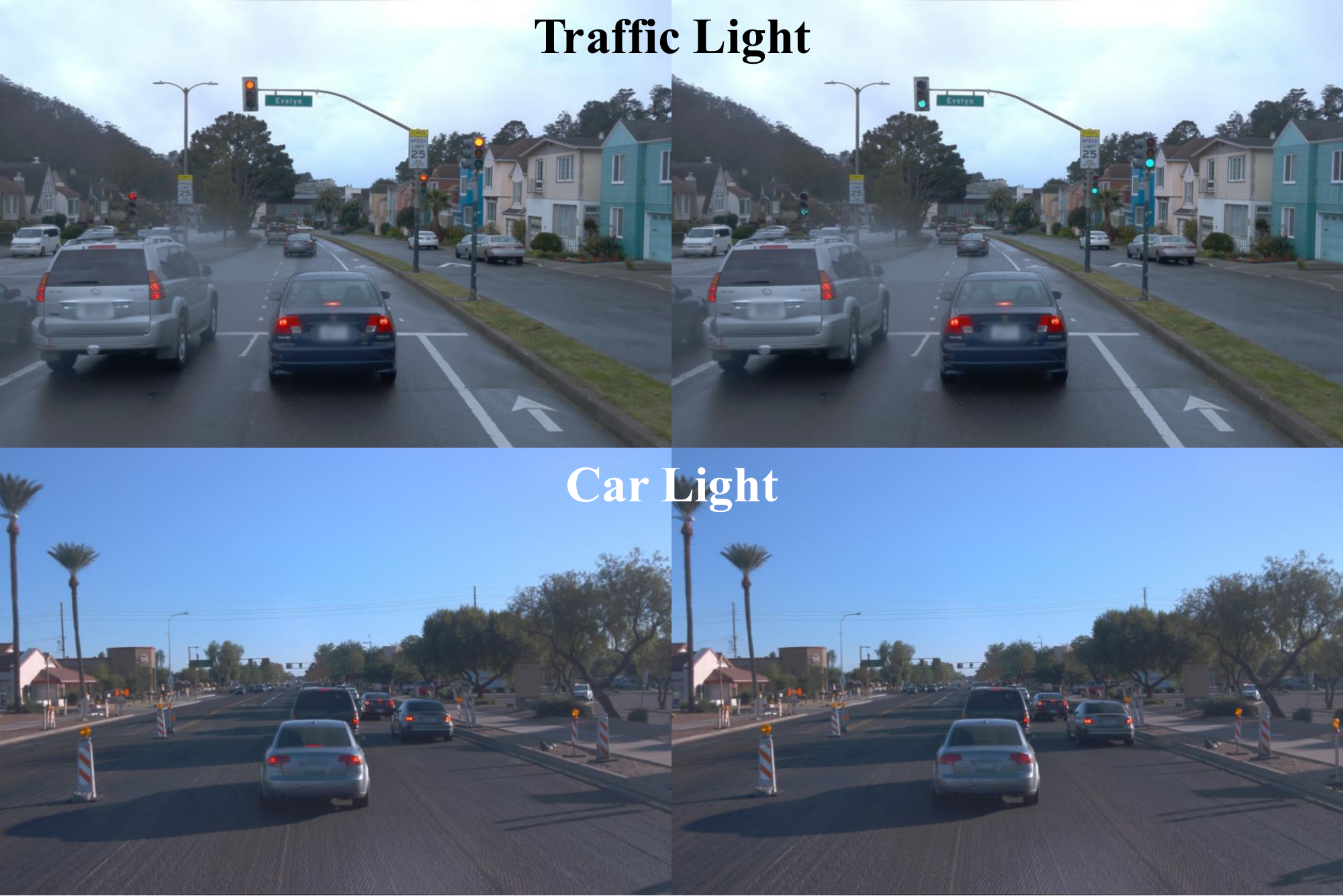}
        \vspace{-6mm}
        \caption{Model light variation using 4DSH.}
        \label{fig:light}
    \end{minipage}
    \vspace{-2mm}
\end{figure*}

\subsection{Rotation}
For rigid dynamics, the z-axis orientation remains fixed, and rotation occurs only within the xy-plane. Therefore, we rotate the Gaussian primitives based on the tangent direction of the trajectory projected onto the xy-plane rather than interpolate the quaternions. While this rotation modeling may not be ideal for non-rigid dynamics, experiments in the supplementary material show that it still effectively models the motions of non-rigid ones.

\subsection{Hyperparameter Settings}
The loss weights are tuned at a single sequence on Waymo~\cite{sun2020scalability} and applied directly to other sequences on Waymo and even nuPlan~\cite{karnchanachari2024towards}. The consistently reliable reconstruction performance across different sequences and datasets demonstrates the robustness of our method to hyperparameter settings, with no need for re-tuning.

\section{Reconstruction of complex motions}
Compared with fixed offset which can only model rigid objects, our \bz{} offset is suitable for both rigid and non-rigid ones, achieving consistent and reliable reconstruction performance across diverse scenarios. 
For modeling the motion of deformable objects, the RGB loss plays a dominant role in capturing shape deformations, while the $\mathcal{L}_{icc}$ suppresses floaters. 
As illustrated in~\Cref{fig:ped_recon} and the video \textbf{pedestrian.mp4}, our method achieves high reconstruction accuracy for the upper body of pedestrians, while certain artifacts may appear in the leg region. This limitation arises from the inherent smoothness of \bz{} curves, which struggle to capture high-frequency, discontinuous motions such as leg swinging during walking.
However, semantic segmentation~\cite{kirillov2023segany} of the rendered images demonstrates that pedestrians are well segmented, ensuring the correctness of visual perception in closed-loop autonomous driving simulations. Furthermore, since our method and OmniRe~\cite{chen2024omnire} focus on different aspects, for applications requiring high-fidelity pedestrian reconstruction, we can integrate pedestrian module in OmniRe~\cite{chen2024omnire} to obtain improved results.

\section{More results}
Our method achieves high-quality reconstruction for both static and dynamic components, particularly demonstrating a significant advantage over state-of-the-art approaches in dynamic element reconstruction. By effectively capturing dynamic elements, our approach also excels in novel view synthesis at unseen timestamps, ensuring accurate and temporally consistent rendering.
See more qualitative results in the supplementary video \textbf{B\'ezierGS.mp4}.

\noindent{\bf Computational Cost.} As shown in~\Cref{tab:cost tab}, our approach attains superior NVS quality while maintaining the training and rendering efficiency of 3DGS.
\begin{table}[h]
\centering
\footnotesize
\renewcommand{\arraystretch}{0.78}
\resizebox{\linewidth}{!}{
\begin{tabular}{@{}lcccc@{}}
\toprule
\textit{RTX A6000} & GPU Mem. (GB)$\downarrow$ & Training Time (min)$\downarrow$ & FPS$\uparrow$ & PSNR $\uparrow$ \\
\midrule
DeformableGS~\cite{yang2024deformable}  & 22.7 & 123.1 & 4.9  & 29.52 \\
HUGS~\cite{zhou2024hugs}                & 8.8  & 61.4  & 46.1 & 29.34 \\
Street Gaussians~\cite{yan2024street}   & 6.1  & 71.6  & 67.8 & 28.92 \\
OmniRe~\cite{chen2024omnire}            & 12.4 & 103.5 & 58.3 & 29.41 \\
PVG~\cite{chen2023periodic}             & 11.9 & 33.6  & 50.8 & 29.64 \\
\midrule
\textbf{\model{} (Ours)} & 10.7 & 48.6 & \textbf{88.7} & \textbf{31.51} \\
\bottomrule
\end{tabular}
}
\vspace{-2mm}
\caption{\textbf{Computational Cost} on our Waymo sequences.}
\vspace{-3mm}
\label{tab:cost tab}
\end{table}

\noindent{\bf Accurate Separation.} Our method enables effective separation of foreground and background components, including the removal of rigid and non-rigid objects. 
This further validates the robustness and generality of our approach in handling complex dynamic scenes (\Cref{fig:teaser}(b) and \Cref{fig:remove_ped}).

\section{Limitations}

We still have a few limitations for future improvement:
First, our approach relies to some extent on the accuracy of the segmentation model, and low accuracy during training may lead to misinterpretations. 
Additionally, our method does not explicitly model lighting variations, which limits its ability to simulate complex lighting effects and shadow changes in dynamic environments. This challenge can be mitigated to some extent by employing the 4DSH approach introduced in Street Gaussians~\cite{yan2024street}, as shown in~\Cref{fig:light}.

\end{document}